\documentclass[conference]{IEEEtran}
\IEEEoverridecommandlockouts
\usepackage{cite}
\usepackage{amsmath,amssymb,amsfonts}
\usepackage{graphicx}
\usepackage{textcomp}
\usepackage{mathtools}
\usepackage{graphicx}
\usepackage{booktabs}
\usepackage{algorithm}
\usepackage{algpseudocode}
\usepackage{adjustbox}
\usepackage{amssymb}
\usepackage{amsmath}
\usepackage{arydshln}
\usepackage{tabularray}
\usepackage{color, colortbl}
\usepackage[table]{xcolor} % For \rowcolor
\usepackage{array}
\usepackage{float}
\def\BibTeX{{\rm B\kern-.05em{\sc i\kern-.025em b}\kern-.08em
    T\kern-.1667em\lower.7ex\hbox{E}\kern-.125emX}}

\makeatletter
\newcommand{\linebreakand}{%
  \end{@IEEEauthorhalign}
  \hfill\mbox{}\par
  \mbox{}\hfill\begin{@IEEEauthorhalign}
}
\makeatother
\usepackage{hyperref}

\begin{document}

\title{Probing a Vision-Language-Action Model for Symbolic States and Integration into a Cognitive Architecture\\}

\author{
\IEEEauthorblockN{Hong Lu\textsuperscript{*}}
\IEEEauthorblockA{\textit{Department of Computer Science} \\
\textit{Tufts University}\\
Medford, MA, USA \\
Hlu07@tufts.edu}
\and
\IEEEauthorblockN{Hengxu Li\textsuperscript{*}}
\IEEEauthorblockA{\textit{Department of Computer Science}\\
\textit{Tufts University}\\
Medford, MA, USA \\
Hengxu.Li@tufts.edu}
\and
\IEEEauthorblockN{Prithviraj Singh Shahani\textsuperscript{*}}
\IEEEauthorblockA{\textit{Department of Computer Science} \\
\textit{Tufts University}\\
Medford, MA, USA \\
Prithviraj\_Singh.Shahani@tufts.edu}
\linebreakand
\IEEEauthorblockN{Stephanie Herbers}
\IEEEauthorblockA{\textit{Department of Computer Science} \\
\textit{Tufts University}\\
Medford, MA, USA \\
Stephanie.Herbers@tufts.edu}
\and
\IEEEauthorblockN{Matthias Scheutz}
\IEEEauthorblockA{\textit{Department of Computer Science} \\
\textit{Tufts University}\\
Medford, MA, USA \\
Matthias.Scheutz@tufts.edu}
}

\maketitle

\begin{abstract}
Vision-language-action (VLA) models hold promise as generalist robotics solutions by translating visual and linguistic inputs into robot actions, yet they lack reliability due to their black-box nature and sensitivity to environmental changes. In contrast, cognitive architectures (CA) excel in symbolic reasoning and state monitoring but are constrained by rigid predefined execution. This work bridges these approaches by probing OpenVLA’s hidden layers to uncover symbolic representations of object properties, relations, and action states, enabling integration with a CA for enhanced interpretability and robustness. Through experiments on LIBERO-spatial pick-and-place tasks, we analyze the encoding of symbolic states across different layers of OpenVLA's Llama backbone. Our probing results show consistently high accuracies ($>0.90$) for both object and action states across most layers, though contrary to our hypotheses, we did not observe the expected pattern of object states being encoded earlier than action states. We demonstrate an integrated DIARC-OpenVLA system that leverages these symbolic representations for real-time state monitoring, laying the foundation for more interpretable and reliable robotic manipulation.
\end{abstract}

\begin{IEEEkeywords}
Vision Language Action Model, Symbolic States, Cognitive Architectures, Robotics
\end{IEEEkeywords}

\textsuperscript{*}These authors contributed equally.

\section{Introduction}
A vision-language-action (VLA) model is a type of foundation model for robotics that takes in images and language commands as input and directly outputs robot actions \cite{team_octo_2024, kim_openvla_2024}. VLAs show promise in providing generalist robot policies across different scenarios and robotic platforms \cite{team_octo_2024}. Recently, OpenVLA has emerged as a significant open-source VLA model, built on a Llama 2 language model backbone combined with a visual encoder that fuses pretrained features. Despite using only 7B parameters (7x fewer than comparable models), OpenVLA has demonstrated strong generalization capabilities across diverse manipulation tasks through its training on nearly one million real-world robot demonstrations \cite{kim_openvla_2024}.

However, recent evaluation of VLAs shows that they struggle with changes in environmental factors such as camera poses, lighting conditions, and the presence of unseen objects \cite{wang_towards_2024}. VLAs also lack reliability, particularly because of their opaque, black-box nature, which makes their internal workings challenging to interpret. 

On the other hand, traditional Cognitive Architectures (CA), also known as symbolic architectures, excel in dependable, symbol-based reasoning but are constrained by their reliance on predefined rules and coded policy execution \cite{gudivada_chapter_2016, aldinhas_ferreira_overview_2019}. Ideally, a CA could harness the versatility of generalist robotic policies and the multimodal capabilities offered by VLAs while maintaining vigilance over dynamic environmental changes during execution in safety-critical applications such as robotic manipulations. 

% In this work, we investigate how OpenVLA encodes state representations through probing. We develop a VLA component in the Distributed Integrated Cognition Affect and Reflection (DIARC) Architecture \cite{aldinhas_ferreira_overview_2019} to demonstrate the first step toward integrating a CA and a VLA.

In this work, we investigate whether and how OpenVLA encodes symbolic representations in its activation space through probing experiments. Our investigation aims to answer the following research questions:

\begin{itemize}
    \item RQ1: To what extent can we decode object properties and relations (e.g., spatial relationships between objects) from OpenVLA's hidden layer activations?
    \item RQ2: Can we extract action-related concepts (e.g., grasp states, movement targets) from the model's activation patterns, and how do these compare to object-level representations?
\end{itemize}

\begin{figure*}[th]
    \centering 
    \includegraphics[width=1.0\linewidth]{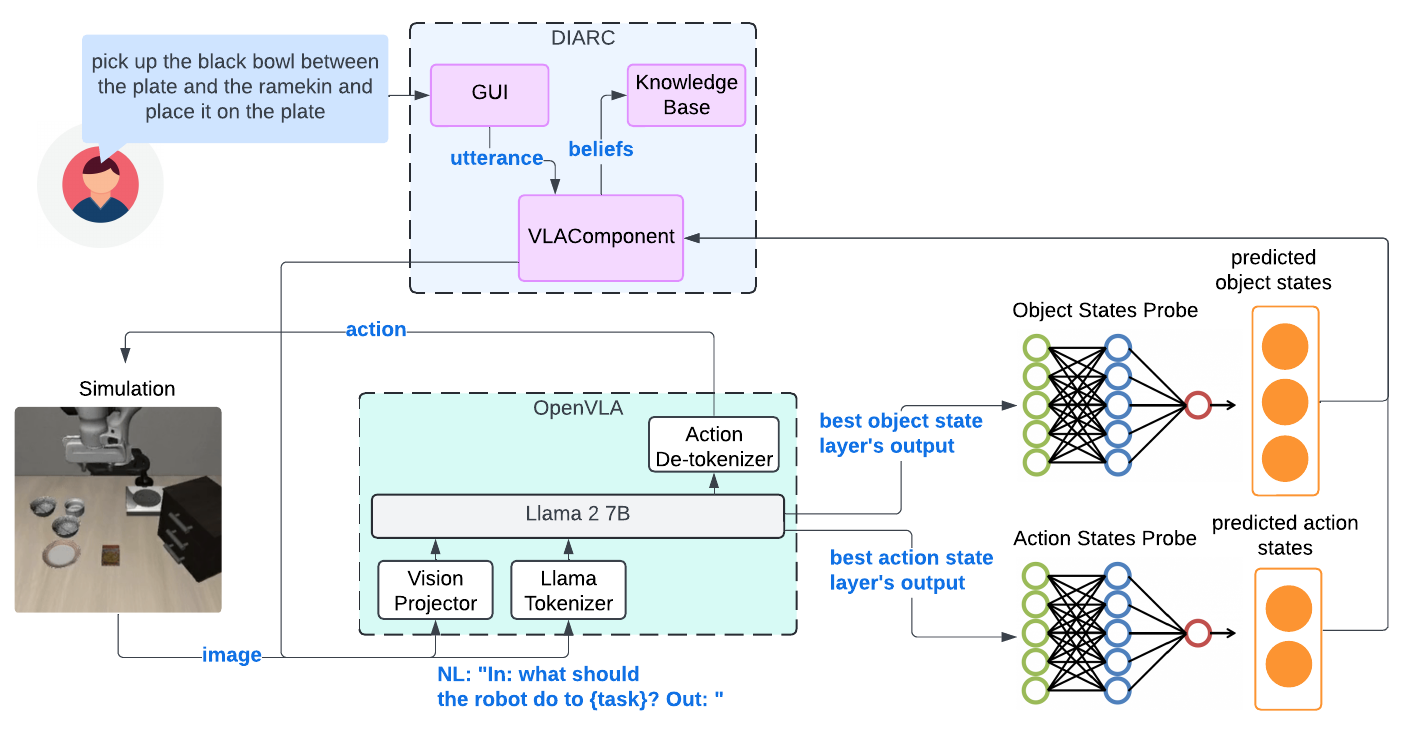}
    \caption{\textbf{The DIARC - VLA - Probe System}. The user selects a natural language command in DIARC's Graphical User Interface (GUI). The VLAComponent in DIARC sends this command to OpenVLA. The probes receive two hidden layers' activations in OpenVLA's Llama backbone that encode the most object state and action state information respectively. The two best hidden layers are identified through the probing experiment described in Section~\ref{sec:probing-experiment}. The probes predict the object state and the action state based on the hidden layers' activations at each timestep. The VLAComponent in DIARC updates DIARC's beliefs based on the predicted object state and action state.}
    \label{fig:diarc-vla}
\end{figure*}

To answer these questions, we train linear probes on different layers of OpenVLA to predict symbolic states during manipulation tasks. Based on prior work in language model probing \cite{chen_is_2024}\cite{touvron_llama_2023}, we hypothesize that:
\begin{itemize}
    \item H1: Object states are primarily encoded in \emph{earlier} layers, as these may encode basic visual and spatial properties.
    \item H2: Action-related concepts are encoded in \emph{later} layers, where visual and language information has been integrated for action planning
\end{itemize}
To test these hypotheses, we conduct a probing experiment in which we examine \emph{all 33 hidden layers} of OpenVLA for their capacity to predict object and action states. Figure~\ref{fig:diarc-vla} shows how we integrate the best-performing layers (for object vs.\ action states) into DIARC for real-time symbolic monitoring of the model's internal state. 
% Through experiments on pick-and-place tasks, we find high probe accuracy ($>0.90$) across most layers for both object and action states, though the hypothesized layer-wise patterns did not emerge. This work provides insights into VLA representations and demonstrates a practical approach for bridging neural and symbolic systems in robotics.

\begin{figure*}[th]
    \centering 
    \includegraphics[width=1.0\linewidth]{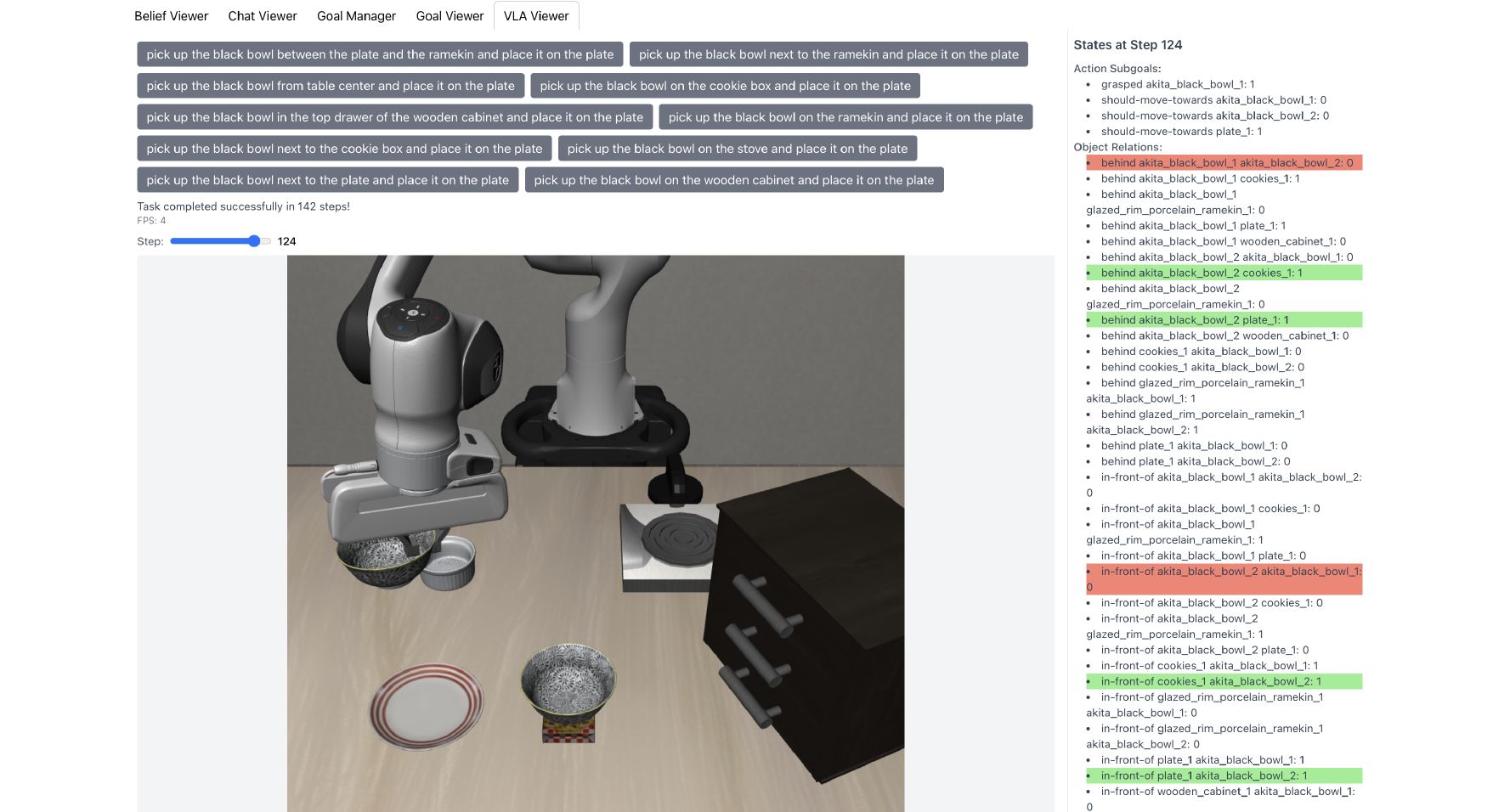}
    \caption{
    \textbf{DIARC--OpenVLA GUI.}
    The left-hand pane displays the real-time camera feed (updated at 5--10\,Hz), showing the robot’s manipulation progress. The right-hand pane color-codes each predicted symbolic state (green for newly activated, red for deactivated), letting users quickly verify whether OpenVLA’s internal representation matches the environment. After task completion, a timeline slider appears, allowing the user to revisit earlier steps’ images and states for deeper analysis.
    }
    \label{fig:ui-screenshot}
\end{figure*}

\section{Related Work}
\subsection{Cognitive Architecture - Foundational Model Integration}
Existing works typically explore how CAs can be integrated with large language models (LLMs). For example, Wu et al. investigate whether the Llama-2 13B model encodes features that can predict expert decisions in a decision making task by training a linear classifier on the Llama model's last contextual embeddings to predict ACT-R's expert decision when the model is given ACT-R's strings of decision making traces as input; they further examine whether ACT-R's knowledge can be injected into the Llama model by fine-tuning a Llama-classifier system on ACT-R's expert decisions \cite{wu_cognitive_2024}. Bajaj et al. enhance ACT-R's analogical reasoning capabilities by building a natural language processing pipeline to automatically extract key entities, relationships, and attributes \cite{bajaj_generating_2023}. Once these key elements have been extracted from unstructured text, an LLM is prompted to convert the unstructured text into a structured format based on its key elements. ACT-R can utilize the structured knowledge for reasoning tasks downstream, significantly reducing the need for manual knowledge engineering. While Bajaj et al. propose to use LLMs to transform unstructured text into structured knowledge, Kirk et al. explore ways in which LLMs can be leveraged as a knowledge source for the task knowledge needed for successful task planning downstream \cite{kirk_exploiting_2023}. They propose three approaches to knowledge extraction: indirect extraction in which the LLM's reponses are placed in a knowledge store that the cognitive agent accesses, direct extraction in which the agent directly queries the LLM and parses its output for structured knowledge, and direct knowledge encoding in which the LLM creates programs that are run as part of the cognitive agent's task pipeline. 

To the best of our knowledge, no existing work has explored the integration of a CA with a VLA model. Realizing such an integration requires a method to ``probe'' the VLA’s hidden representations for symbolic content. We next discuss relevant literature on foundational model probing, which informs our approach in extracting object and action states from OpenVLA.

\subsection{Foundational Model Probing}
Foundational models encode extensive knowledge derived from their internet-scale training data \cite{bommasani_opportunities_2022}. The increasing popularity of these models has drawn significant attention to the challenges of extracting and evaluating the knowledge they encode. One commonly used approach to evaluate LLMs is prompt-based probing in which an LLM is prompted to fill in the blanks in the prompt \cite{petroni_language_2019}. For example, Alivanistos et al. combine various prompting techniques to probe LLMs for possible objects of a triple where the subject and the relation are given \cite{alivanistos_prompting_2023}, Wang et al. develop a method to automatically search sentences to construct optimal and readable prompts for probing certain knowledge \cite{wang_readprompt_2023}, and Qi et al. propose a probing framework for multimodal LLMs that includes visual prompting, textual prompting, and extra knowledge prompting \cite{qi_what_2023}. While prompt-based probing is intuitive and easy to execute, it lacks the layer-specific precision offered by linear probing. Furthermore, prompt-based probing is not applicable to vision-language-action models as they do not output language tokens. Linear probing on the other hand involves training a linear classifier on top of each frozen layer of a foundational model. Each classifier is tasked with predicting specific knowledge based on the output features of the corresponding frozen layer. For example, Li et al. train semantic probes to predict object properties and relations as they evolve throughout a discourse \cite{li_implicit_2021}. Similarly, Chen et al. use linear probes to evaluate the Llama model family’s performance on higher-order tasks such as reasoning and calculation, comparing probe performance across layers and model sizes \cite{chen_is_2024}. In our work, we extract symbolic representations of state changes similar to the approach described in Li et al \cite{li_implicit_2021} and we evaluate probing accuracies across hidden layers similar to the approach used by Chen et al \cite{chen_is_2024}.

\section{Integrated Vision-Language-Action Model - Cognitive Architecture Overview}
\label{sec:integration-overview}

Figure~\ref{fig:diarc-vla} illustrates the high-level architecture of our DIARC--OpenVLA system.
DIARC provides a cognitive architecture that manages symbolic reasoning and user interaction,
while OpenVLA is a continuous policy that takes in images and language instructions to produce
a 7D robot action. Internally, OpenVLA uses a Llama 2 7B backbone \cite{touvron_llama_2023}, which consists of
32 transformer blocks (often referred to as “layers”) plus an initial embedding layer,
yielding 33 distinct hidden states when indexed from 0 to 32 at runtime.
Each hidden state is a 4096-dimensional vector.
At a conceptual level, we combine these by:
(1) routing user commands through DIARC to OpenVLA, 
(2) running OpenVLA in a LIBERO simulation environment
to generate actions and extract hidden-layer embeddings,
and (3) mapping those embeddings to symbolic states for DIARC’s belief store.
This pipeline leverages the expressiveness of a vision-language-action model
while maintaining the reliability of a symbolic architecture.
Subsection~\ref{subsec:diarc-integration} details our real-time implementation,
including the WebSocket interface and a React-based UI for visualization.

% \subsection{DIARC VLA Component}
% \textbf{add something here to describe the DIARC VLA UI}
% The VLAComponent receives two arrays containing the probes prediction outputs for the object and actions states containing zeros and ones with ones indicating true and zeros indicating false. Then, the VLAComponent converts that information into predicates that is in the form of the data useful to DIARC. Specifically, the predicate format takes the form of \textit{relation(object1, object2)} or \textit{property(object)} for object states and \textit{action(object)} for actions states. These predicates form the beliefs that DIARC holds. 

\subsection{DIARC--OpenVLA Integration}
\label{subsec:diarc-integration}

The DIARC--OpenVLA integration bridges OpenVLA’s continuous policy outputs and hidden-layer embeddings
with DIARC’s symbolic reasoning modules. Our approach requires minimal modification to OpenVLA itself:
we intercept each inference call to extract the relevant hidden-layer activations and feed them to
trained linear probes for symbolic state prediction, then pass those states back to DIARC in an
automated fashion. Figure~\ref{fig:diarc-vla} provides a broad schematic, and
Figure~\ref{fig:ui-screenshot} shows the user interface in action.

\paragraph{VLAComponent and Symbolic Predicates.}
At every timestep, OpenVLA predicts a 7D action $\Delta x, \Delta \theta, \Delta \text{Grip}$ given
the current camera image and the user’s natural-language instruction.
In parallel, we run linear probes on the extracted hidden-layer embeddings
to output arrays of 0/1 labels for object relations and action subgoals
(e.g., \emph{on(bowl, plate)} = 1, \emph{grasped(bowl)} = 0).
These arrays are sent to DIARC’s VLAComponent, which converts them into DIARC’s
symbolic predicate format: \emph{relation(object1, object2)}, \emph{property(object)} for object states,
and \emph{action(object)} for action states. For example, a 1 in \emph{grasped(bowl\_1)} becomes
\emph{grasped(bowl\_1)} in DIARC’s knowledge store. DIARC can then leverage these discrete predicates
to detect inconsistencies (e.g., a bowl cannot be both \emph{on(bowl\_1, plate\_1)} and 
\emph{inside(bowl\_1, drawer\_1)} at the same time), verify subgoals, or track overall task progress.

\paragraph{WebSocket Server and Real-Time Flow.}
We implement a lightweight WebSocket server to provide real-time communication among OpenVLA,
the LIBERO simulator, DIARC, and a React UI:
\begin{enumerate}
    \item \textbf{User Task.} The user selects a pick-and-place command in DIARC’s GUI
    (e.g., ``pick up the black bowl between the plate and the ramekin...'').
    DIARC sends this instruction via WebSocket to our server.
    \item \textbf{Environment Step.} The server runs the LIBERO-spatial simulator,
    retrieving the latest camera frame for input to OpenVLA’s policy.
    OpenVLA returns a 7D action, which the simulator executes.
    \item \textbf{Probe Inference.} Simultaneously, the server extracts the hidden-layer embedding
    from OpenVLA, feeds it to our linear probes, and obtains predicted symbolic states
    (object relations, subgoals, etc.).
    \item \textbf{Streaming Back.} Finally, the server encodes the current camera image in base64
    and bundles it with the predicted symbolic states plus the timestep index.
    This data is streamed back over the WebSocket to DIARC and the React UI.
\end{enumerate}

\paragraph{React UI and Timeline Scrubbing.}
Figure~\ref{fig:ui-screenshot} shows our React-based interface. While the task runs,
the UI continuously displays:
\begin{itemize}
    \item A \emph{live camera feed} (at $\sim$5\,Hz) pinned at the top-left, showing the robot’s current manipulation.
    \item A \emph{symbolic states} panel on the right, color-coding newly activated or deactivated predicates
    (e.g., green if \emph{on-table(bowl\_1)} flips from 0 to 1).
    \item A \emph{timeline slider} becomes available once the task completes, letting the user “scrub” back
    through each timestep’s image and states to analyze the model’s evolution over time.
\end{itemize}
This design allows operators or domain experts to confirm that the predicted states mirror
actual environment changes (e.g., verifying \emph{on(bowl,plate)=1} precisely when
the bowl is placed). Meanwhile, DIARC receives the same 0/1 states as symbolic predicates,
enabling high-level logic or safety checks without manually altering the VLA policy. By decoupling the raw policy (OpenVLA) from DIARC’s symbolic reasoning through a WebSocket-based design,
we maintain modularity while allowing step-by-step monitoring of the policy’s internal states.
We thus leverage the \emph{generalist} capabilities of a VLA model and the \emph{reliability}
of a symbolic architecture. This opens the door for future enhancements where DIARC might intervene
upon contradictory states (e.g., an object can’t be both ``on the plate'' and ``in the cabinet'')
or respond to user queries mid-task. In short, our integration unites robust low-level action generation
with high-level interpretability and control.

\subsection{Simulated Pick-and-Place Tasks}
\label{subsec:pickplace-tasks}
LIBERO-spatial is a suite of 10 pick-and-place tasks in the LIBERO simulation environment \cite{liu_libero_2023}. We choose LIBERO-spatial for our OpenVLA evaluation as a LIBERO-spatial finetuned OpenVLA checkpoint is readily available for download. The LIBERO-spatial task suite consists of 10 pick-and-place tasks of the form ``pick up the black bowl \{spatial relations identifier\} and place it on the plate'' where the ``spatial relations identifier'' is filled with a natural language description of the target black bowl's spatial relations to its surrounding objects. For example, Figure~\ref{fig:symbolic_states} shows 4 frames of the OpenVLA performing the ``pick up the black bowl between the plate and the ramekin and place it on the plate'' task. Other LIBERO-spatial pick-and-place tasks include ``pick up the black bowl next to the ramekin and place it on the plate'' and ``pick up the black bowl in the top drawer of the wooden cabinet and place it on the plate''. Since the 10 pick-and-place tasks involve the same objects and the object initial placements remain the same across tasks except for the two black bowls, the natural language description of the target black bowl's spatial relations serves as an identifier.

\section{Probing Experiment}
\label{sec:probing-experiment}
To test our hypotheses, we extract activations from the 33 hidden layers of OpenVLA's 
Llama 2 7B backbone. Each hidden-layer embedding is a 4096-dimensional vector. We then train two probes on each layer's 
activations to predict object states and action states. In total, we train $2 \times 33 = 66$ probes.

An object state involves the following relation predicates: \textit{behind(tabletop-object1, tabletop-object2)}, \textit{in-front-of(tabletop-object1, tabletop-object2)}, \textit{inside(tabletop-object, container)}, \textit{left-of(tabletop-object1, tabletop-object2)}, \textit{on(tabletop-object1, tabletop-object2)}, \textit{on-table(tabletop-object)}, and \textit{right-of(tabletop-object1, tabletop-object2)}, as well as unary object property predicates: \textit{open(container)} and \textit{turned-on(on-off-object)}.

An action state captures the action status predicate \textit{grasped(pickupable-object)} and the action subgoal predicate \textit{should-move-towards(tabletop-object)}. Examples of object states and action states are provided in the Probe Training Data Collection section below. We find combinations of grounded objects to which a predicate is applicable, and we define the object relation atoms as the object relation predicates applied to all combinations of their grounded objects. We define an object state as a complete truth assignment to the object relation atoms and object property atoms. Similarly, we define an action state as a complete truth assignment to the action status atoms and action subgoal atoms. In total, an object state has 224 atoms and an action state has 12 atoms.
\subsection{Probe Training Data Collection}
\begin{figure}
    \centering
    \includegraphics[width=0.8\linewidth]{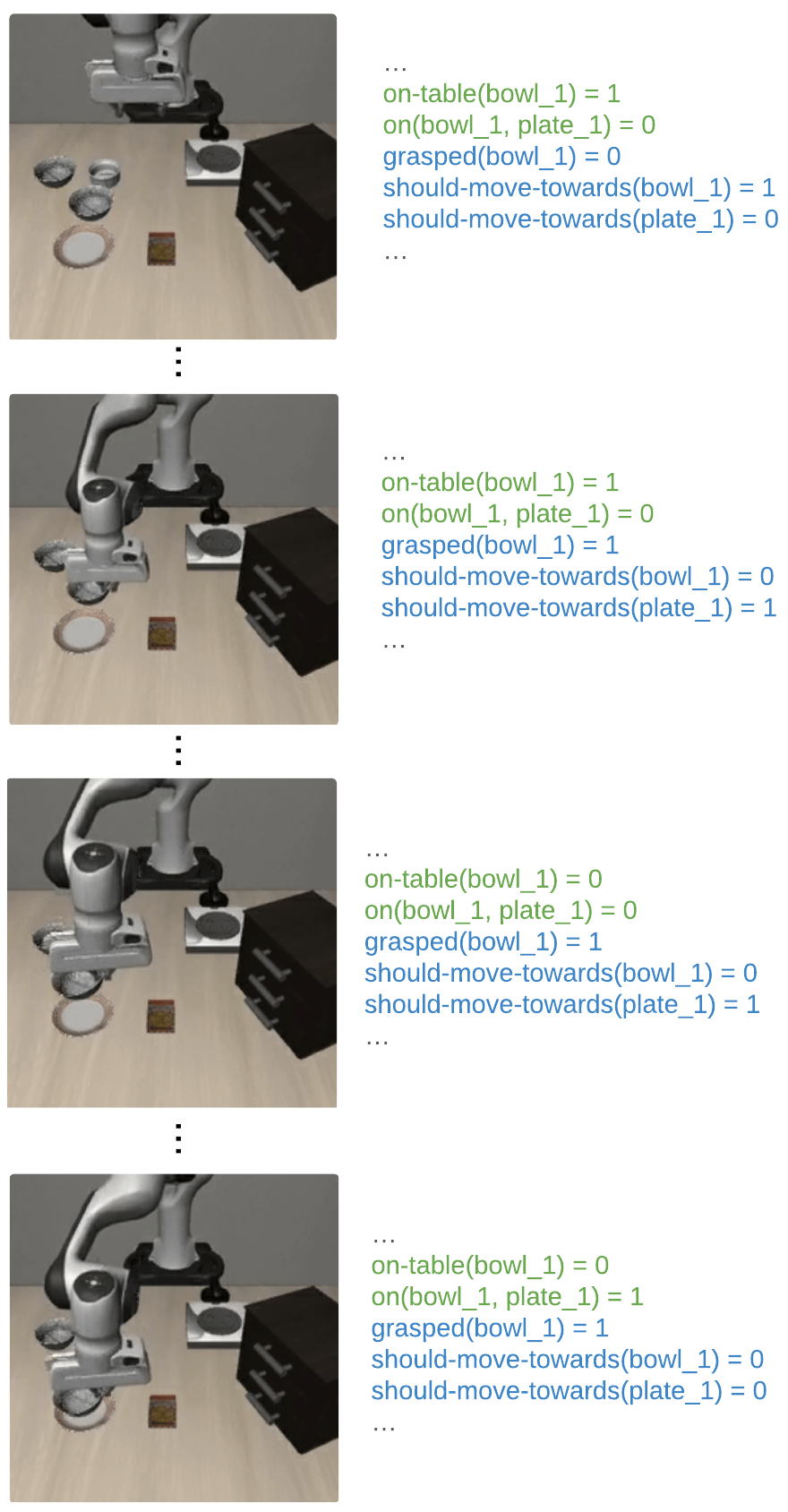}
    \caption{\textbf{Example Labeled Object States and Action States in a Pick-and-Place Trajectory.} Object states are shown in green and action states are shown in blue.}
    \label{fig:symbolic_states}
\end{figure}
To train the probes, we need to collect (hidden layer activation, ground truth state) pairs as training data. To do this, we implement detector functions that detect the truth values for object relations, object properties, action statuses, and action subgoals in the 10 LIBERO-spatial tasks.  Figure~\ref{fig:symbolic_states} shows 4 frames of the OpenVLA performing the ``pick up the black bowl between the plate and the ramekin and place it on the plate'' task with their object state (green) and action state (blue) labeled by the detector functions where a label of 1 represents true and a label of 0 represents false. The 4 frames are points at which critical state changes happen. The first frame (top) is taken at the beginning of an episode where the target black bowl (which happens to be the \textit{bowl\_1} object) is still directly on the table and the robot needs to pick it up. As expected, the \textit{on-table(bowl\_1)} object relation and the \textit{should-move-towards(bowl\_1)} action subgoal are detected to be true. Once the robot arm closes its grippers on the target black bowl in frame 2,  the \textit{grasped(bowl\_1)} action status is detected to be true. In frame 3, the \textit{on-table(bowl\_1)} object relation becomes false as the bowl is lifted off the tabletop. Finally, in frame 4, the \textit{on(bowl\_1, plate\_1)} object relation becomes true, indicating that the task has been successfully completed.

We collect 5 successful episodes per LIBERO-spatial task by repeatedly querying 
OpenVLA until 5 completed episodes are obtained. While the model is queried to predict 
the next robot action at each frame, we also extract and record the activations 
from each hidden layer $\ell \in \{0, \dots, 32\}$. Importantly, at timestep $t$, we pair 
the hidden-layer embedding $\mathbf{h}_t$ with the ground-truth symbolic state 
$\mathbf{y}_t$ at the \emph{same} time, ensuring no temporal mismatch 
(e.g., not using $t+1$ states to label time $t$). 
We then store each pair as $(\mathbf{h}_t, \text{object state}_t)$ or $(\mathbf{h}_t, \text{action state}_t)$.

\begin{figure*}[htbp]
    \centering
    \includegraphics[width=1.0\linewidth]{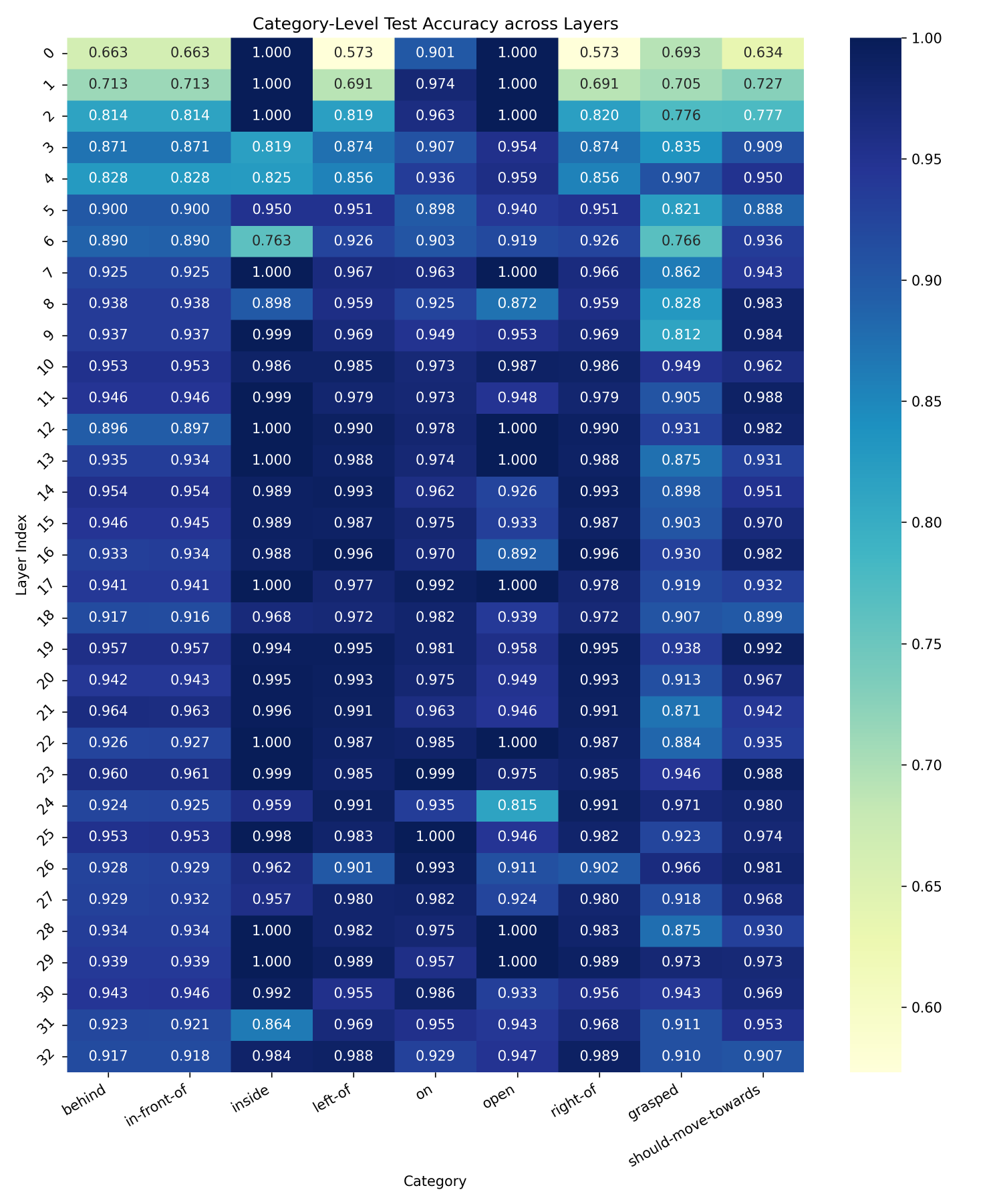}
    \caption{\textbf{Probing Results.} The first seven columns are object state symbols and the last two columns are action state symbols.}
    \label{fig:probing}
\end{figure*}
% \begin{figure*}[th]
%     \centering
%     \includegraphics[width=1.0\linewidth]{probing-heatmap.png}
%     \caption{Probing Results. The first seven columns are object state symbols}
%     \label{fig:probing}
% \end{figure*}
% \subsection{Probe Training Data Preprocessing}
% \textbf{Describe the steps done to clean the transformed data. For example, data labels that rarely change are dropped. Explain the rationale behind dropping them. Some labels are unbalanced. Explain the steps taken to balance them e.g oversampling.}
\subsection{Probe Training Data Preprocessing}

To produce a reliable dataset for training our linear probes, we collect (1) per-timestep embeddings from OpenVLA’s hidden layers and (2) corresponding ground-truth symbolic states (e.g., object relations and action subgoals) from a LIBERO environment. We then apply the following preprocessing steps to ensure that the resulting dataset reflects meaningful, fluctuating states and avoids overfitting or misleading evaluations:

\subsubsection{Episode-Level Splitting for Train/Test}
We first gather a dataset of entire episodes (trajectories) from the environment, storing (embedding, symbolic\_state) pairs at each timestep. To prevent \emph{temporal leakage}---where the model might ``see'' future frames from the same trajectory during training---we split the dataset \emph{by episode}. Specifically, we assign each entire episode to either the training or the test set. This ensures that no frames from the same trajectory end up in both sets, so the probe’s evaluation genuinely measures out-of-episode generalization.

\subsubsection{Filtering Near-Constant Labels}
Some symbolic states never change (or change only once) across our collected episodes, offering little to no discriminative signal. They can also artificially inflate accuracy if trivially predicted (e.g., always zero). We measure each label’s frequency of being `1' across all training frames and remove any label whose frequency is below $1\%$ or above $99\%$. This pruning discards \emph{near-constant} labels, ensuring that the remaining labels reflect genuinely variable states that are amenable to learning.

\subsubsection{No Class Balancing}
While class imbalance is a concern for certain tasks, we do not perform synthetic oversampling or under-sampling in this work. In practice, removing near-constant labels already eliminates the most extreme forms of imbalance, so we find that further balancing is not strictly necessary. Nevertheless, class balancing remains an option for future improvements if highly skewed distributions re-emerge with other data.

\subsubsection{No Feature Standardization}
We load the embeddings as raw floating-point vectors from OpenVLA’s hidden layers and do \emph{not} apply z-score normalization or any other scaling. Although standardizing embeddings can help in some contexts, we observe that the linear probe converges well without it---likely because the embeddings remain within moderate numeric ranges. We thus preserve a simpler pipeline, though we note that feature standardization remains a viable tweak for additional robustness.

\subsubsection{Resulting Training/Testing Dataset}
After these steps, we obtain a final dataset whose labels reflect genuinely changing states, each entire episode is restricted to one split (train or test), and no artificially inflated metrics arise from trivial or constant conditions. We train the linear probe on the processed training subset and evaluate it on disjoint test episodes, confident that the reported performance tracks the probe’s capacity to decode meaningful state information from the model’s representations.

\subsubsection{Summary}
By removing near-constant labels, splitting entire episodes for train/test, and keeping the embeddings unmodified, we produce a dataset that highlights OpenVLA’s genuine internal representations of symbolic states. This pipeline avoids temporal leakage and trivial labels, thereby enabling a fair and interpretable evaluation of the probe’s performance.

% \subsection{Probe Training and Evaluation}
% \textbf{Describe the architecture of the probe and cite relevant sources. Describe how we train and evaluate the probe. For example, how we split train and test data, how we measure accuracy, how we averaged the accuracies for the same predicate across atoms e.g. the accuracies for on(bowl\_1, plate\_1), on(plate\_1, wooden\_cabinet\_1) etc. are averaged to simply be ``on''}
% Prior work has probed language models

\subsection{Probe Training and Evaluation}
Our probing methodology builds on recent work investigating internal representations in language models \cite{li_implicit_2021} and multimodal embeddings \cite{dahlgren-lindstrom-etal-2020-probing}. Inspired by these approaches, we implement a linear probe that maps from the model's internal representations to ground truth environment states. However, rather than using single-label classification which would face combinatorial explosion with growing numbers of states, we extend this to multi-label classification where each state variable can be predicted independently.

Formally, for a given layer's activation vector $\mathbf{h} \in \mathbb{R}^d$, our probe learns a mapping to binary predictions $\hat{\mathbf{y}} \in [0,1]^n$ where $n$ is the number of tracked symbolic states:

\begin{equation}
\hat{\mathbf{y}} = \sigma(\mathbf{W}\mathbf{h} + \mathbf{b})
\end{equation}

where $\mathbf{W} \in \mathbb{R}^{n \times d}$ and $\mathbf{b} \in \mathbb{R}^n$ are learned parameters and $\sigma$ is the sigmoid activation function. Each element of $\hat{\mathbf{y}}$ corresponds to a binary prediction about a specific ground atom (e.g., ``on(bowl\_1, plate\_1)'' or ``in(bowl\_1, top\_drawer''). We train using binary cross-entropy loss with the Adam optimizer.

For evaluation, we compute averaged accuracies per predicate type. For a predicate like ``on'', we average the accuracies across all specific instances of that predicate that we track - for example, if we track both ``on(bowl\_1, plate\_1)'' and ``on(plate\_1, table\_1)'', we would average their individual prediction accuracies to get the overall accuracy for the ``on'' predicate:

\begin{equation}
\text{acc}(\text{pred}) = \frac{1}{N_\text{pred}} \sum_{i=1}^{N_\text{pred}} \text{acc}(i)
\end{equation}

where $N_\text{pred}$ is the number of tracked instances of that predicate, and $\text{acc}(i)$ is the binary prediction accuracy for the $i$th instance.

The probe results, visualized as a heatmap across layers and predicates, reveal consistently high accuracies across later layers suggesting robust encoding of symbolic state information. However, the first layer shows notably lower performance, aligning with its expected role in encoding lower-level features rather than high-level semantic relationships.

\section{Results and Discussion}
% present and discuss your results and draw any conclusion from them for the future of cognitive architecture and FMs
Figure~\ref{fig:probing} shows a heatmap of the object state probes' accuracies and the action state probes' accuracies across all 33 layers. The first 7 categories are object relations and properties whereas the last two categories are action status and subgoal. Note that the object relation \textit{on-table(tabletop-object)} and the object property \textit{turned-on(on-off-object)} are dropped from training due to low variance in the training data-most objects remain on-table and the \textit{stove\_1} object (the only on-off-object) remains off, therefore their corresponding atoms never change truth values. 

The accuracies are above 0.90 for most layers, indicating that the OpenVLA indeed encodes some object relation, object property, action status, and action subgoal features.  The layer 0 probes perform significantly worse across all categories. This is not surprising as the first Llama layer probably only encodes very low-level semantic features such as syntactic relations and not the high level visual-semantic features such as object relations. We do not observe the hypothesized pattern of higher object state accuracies in earlier layer probes versus that of later layer probes. This does not support our hypothesis 1 and hypothesis 2. We recognize that the training data we use are not diverse enough in terms of the variation in object states and action states. Specifically, the objects in the 10 simulated LIBERO-spatial tasks have the same placements across tasks except for the two black bowls. As a result, most of the object relations remain unchanged across tasks, significantly reducing the variation in object states. Furthermore, the robot always picks one of the two black bowls, and the place target is always the plate, significantly reducing the variation in action states. These two factors combined significantly reduce the difficulty of the linear classification task that the probes are trained to do, leading to high accuracies across layers and categories, potentially washing out the layer-wise object state versus action state difference we expected to observe. More and better data is needed to test our hypotheses.

As future work, we plan to scale the probing experiment up by collecting more diverse data from tasks that involve variable objects, variable object layouts, and variable goals. We believe that extracting symbolic information from VLAs opens the door to the integration of CA and VLA and we demonstrate an integrated CA-VLA system with this work. In the future, we hope to explore how the reasoning capabilities of the CA can enhance or monitor the performance of the VLA.
\bibliographystyle{IEEEtran}
%\bibliographystyle{unsrtnat}
% \balance
\bibliography{main}

\clearpage

\end{document}